# Semantic Video Segmentation: A Review on Recent Approaches

Mohammad Hajizadeh Saffar[1]. Mohsen Fayyaz[2]. Mohammad Sabokrou[3]. Mahmood Fathy[1]

[1]School of Computer Engineering, Iran University of Science & Technology, Tehran, Iran

[2]Faculty of Computer Science, University of Bonn, Bonn, Germany

[3]Computer Science department, Fundamental Sciences (IPM), Tehran, Iran

## Abstract

This paper gives an overview on semantic segmentation consists of an explanation of this field, it's status and relation with other vision fundamental tasks, different datasets and common evaluation parameters that have been used by researchers. This survey also includes an overall review on a variety of recent approaches (RDF, MRF, CRF, etc.) and their advantages and challenges and shows the superiority of CNN-based semantic segmentation systems on CamVid and NYUDv2 datasets. In addition, some areas that is ideal for future work have mentioned.

## 1. Introduction

Semantic segmentation, which means segmentation of all objects of the scene and classifying them based on concept, is one of the fundamental tasks in the field of machine vision. This field is important because it can be considered as a substantial preprocessing for others tasks, including object detection, scene understanding and scene parsing. Semantic segmentation analyzes and classifies the concept and nature of objects, as well as recognizing them and their shape in the scene [1]. For this reason, it can be consisted of three basic steps of object detection, shape recognition and classification [2]. Each of these three steps requires research to improve efficiency, and many researchers have studied one or more of the steps and introduced improvements in this regard.

Obviously, these three steps cannot be considered separate from each other, and the accuracy and efficiency of each of them has a direct impact on next steps. For example, a system which has error in object detection, most probably cannot achieve high performance in shape recognition. Even if the system can recognize the shape with the help of ancillary tools, it will be in trouble in the third step, i.e.

classification. For this reason, a procedure that examines all three steps together and can enhance system performance in a balanced manner, is of great importance. It is natural that a large part of this task in based on the model that should be trained for the system. Many parameters have a direct impact on the quality of the proposed model, including the quality of the dataset used, extracted features, learning and classification methods.

One of the problems that have existed since the introduction of machine learning, as a major field of designing and providing various functional systems, is the lack of large, complete and reference datasets. Of course, this problem has been solved to some extent over time and reference datasets with standard procedures of training and testing have been well developed. But they are still deficient in terms of volume and comprehensiveness of the available categories. Most recently, with a good growth that has been achieved in the field of labeled dataset, semantic segmentation has also attracted the attention of machine vision researchers. Given that the growth and development of the datasets with pixel-level labeling is costly in terms of time and may include human errors [3], semi-supervised and weakly supervised methods benefit from their own advantages and possess their own place among researchers. Besides their different benefits, it is also necessary to note that datasets with video-level labeling cause problem in location detection and segmentation, and therefore some other methods should be adopted to maintain system performance, which increases the cost of designing and developing systems based on such datasets [4]. For example, graphical models, such as conditional random fields (CRF), which requires a high volume of training data with high accuracy labeling, are not suitable for working on these datasets [1].

The general attitude of researchers to the field of semantic segmentation is based on classification, but some of them convert it to a clustering problem [5-7]. Although computational requirements of clustering algorithms are partly manageable on images, but the conditions are completely different for videos. Due to the hardware limitations and deficiencies in conjunction with the computational requirements for clustering videos, this task is usually implemented and run in two separate steps, namely the graph construction and the clustering [5]. The graphs provide a natural representation of image or video sequence so that its edges include the available spatio-temporal structure and their transitivity property provide inferences in the long term [7].

As mentioned above, the semantic segmentation is usually divided into three steps including object detection, shape recognition and classification. One of the common methods for step of object detection is the use of a large number of candidate Boxes and then sorting them. In this case, the quality of classification depends on the quality of system assumptions and possibilities [8]. Semantic part segmentation is an example of candidate box based systems to detect objects in the scene [9]. Of course, another methods are also available for object detection and other steps of designing a semantic segmentation system, which will be mentioned in the following sections.

More than two decades have passed since the introduction of semantic segmentation field as one of the basic pre-processing steps for machine vision based systems, but due to extensive changes in the appearance of objects such as angle and direction, size and scale, blurring and reduced quality, camouflage of objects in the environment, objects overlap, etc. Major challenges have remained in this filed [10]. However, recently with the introduction of deep neural networks (DNNs) and other related concepts, a significant proportion of these problems have been solved, and systems based on these concepts have shown significant efficiency. In the past, segmentation concept involved a combination of semantic and spatial properties defined by the system designer. But with the introduction of deep networks, as one of the strongest fields of machine learning and modeling, these features, along with a deep layer hierarchy and in a non-linear space, created a combination which is more efficient than the old systems. Also according to the studies and obtained results, it is known that the efficiency of these networks enhances by increasing of the networks depth. Over time, feature extraction with the help of deep networks have been taken into consideration, and the use of frameworks provided by very large datasets such as ImageNet [11] have extended.

Along with a variety of tools and methods proposed in the field of deep networks, convolutional neural networks (CNNs) have shown an excellent performance in different fields of image and video such as image classification, object detection, visual tracking and action recognition. The strength of CNN networks in feature representation is much better than other methods, which has led researchers to use it for structural problems such as semantic segmentation and object pose estimation [12]. After introduction of deconvoutional neural networks (DeConvNets) in [13], researchers attempted to change the structure of CNN and converted its architecture into a fully convolutional network (FCN), so that they could obtain to a large map of the labels for the overall image by classification of each small region of the image. Also, by implementation of a linear interpolation, they could carry out the deconvolution process and achieve pixel labels of the image [12].

In this paper, first we review the common datasets related to the field of semantic video segmentation segments and different methods used to evaluate efficiency and accuracy of the system. Then, we will review the works done by other researchers and the methods implemented by them for each of the steps involved in the development of a semantic segmentation system. Also, we well explain a brief introduction to the methodology of convolutional neural networks. In addition, we will review remain challenges and future works in this field.

## 2. Datasets and Evaluation

Approaches need constant testing structure and standard measure parameters to be compared with others. For this reason, some datasets have become standard metric for evaluation of a proposed system accuracy. Also, several quality measures have become common evaluation parameters. In this section, common datasets that have been used in semantic video segmentation will be introduced. Also, most frequently used measures for approach evaluation will be described.

### 2.1. Datasets

Dataset which is selected and used by system designers play a very important role in the quality of the trained model and thereby system performance. Important parameters are involved in selecting dataset, that usually all of them are not exist in a dataset. So, selecting an appropriate dataset for a task can be one of the most challenging work at the beginning of the research process. Among the most important definable parameters for a full dataset, the followings can expressed:

- ❖ Comprehensiveness of dataset in term of its different available categories
- ❖ Input quality and size
- ❖ High proportion of the training and test samples for each of the available categories
- ❖ High accuracy training labels
- ❖ Authenticity and quality of training labels
- ❖ And …

There are various datasets with semantic labels on their video frames. Among them, large differences can be observed in volume of the training and test data, type of training label accuracy based on pixel or video (sometimes referred to as weakly annotate) and different semantic categories. But one of the reasons for the superiority of a dataset on others may be researchers' interest in using it and comparing their works with others. For this reason, some datasets can be considered as original references for a research topic. In the next subsections, a number of mostly used datasets in semantic segmentation will be described.

### 2.1.1. CamVid Dataset

The Cambridge-driving Labeled Video Database (CamVid) is the first collection of videos with object class semantic labels, complete with meta-data [14, 15]. The dataset provides ground-truth labels that associate each pixel with one of 32 semantic classes. In spite of many existing datasets in this area that have been developed using fixed cameras, these series were collected using the cameras installed on a car and moving on city streets. CamVid datase has more than 10 minutes of video with 30Hz quality, which more than 700 frames is labeled manually and has been reviewed and approved by someone else. Some samples of Camvid dataset have been illustrated in Figure 1.

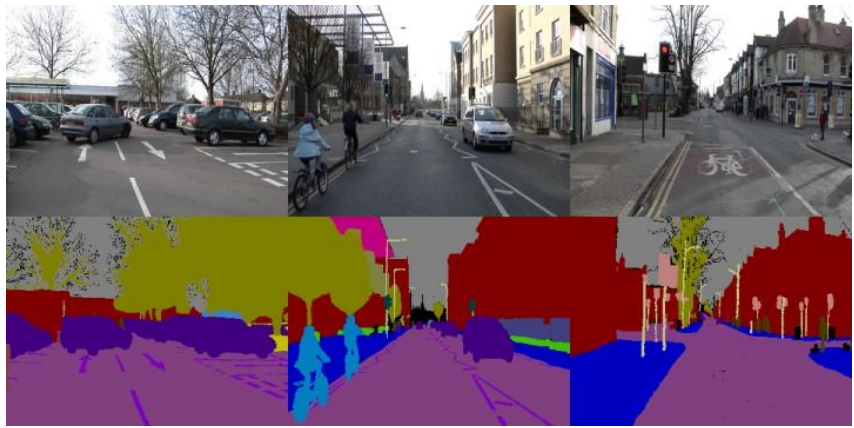

*Figure 1 Sample input frames of CamVid Dataset and it's ground-truth*

### 2.1.2. NYU-Depth V2 Dataset

The NYU-Depth V2 data set is comprised of video sequences from a variety of indoor scenes as recorded by both the RGB and Depth cameras from the Microsoft Kinect [16]. It features 1449 densely labeled pairs of aligned RGB and depth images, 464 new scenes taken from 3 cities and 407,024 new unlabeled frame. This dataset has more than 1000 classes in from 26 different scene types (office room, bedroom, office hallway, living room etc.). Some samples of NYU-Depth dataset have been illustrated in Figure 2.

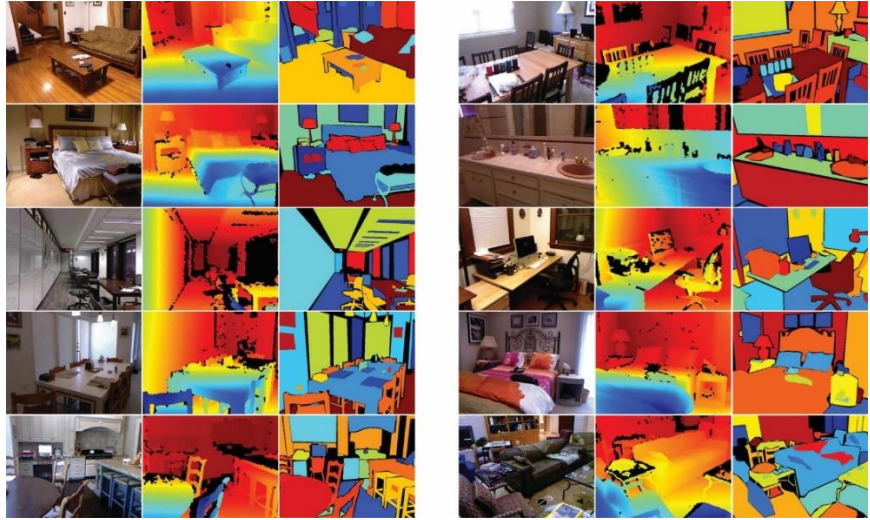

*Figure 2 Sample input frames of NYU-Depth Dataset and it's ground-truth and depth notation*

### 2.1.3. Wild8 Dataset

Liu et al. provide the Wild8 dataset [17] consisting of 100 sequences of weakly supervised videos from 3 documentary series of which 33 sequences are manually labeled with pixel-level ground-truth for evaluation. This dataset includes eight categories (bird, lion, elephant, sky, tree, grass, sand, and water). Wild8 is a multi-class video segmentation dataset and all the sequences are associated with multiple categories. Some samples of Wild8 dataset have been illustrated in Figure 3.

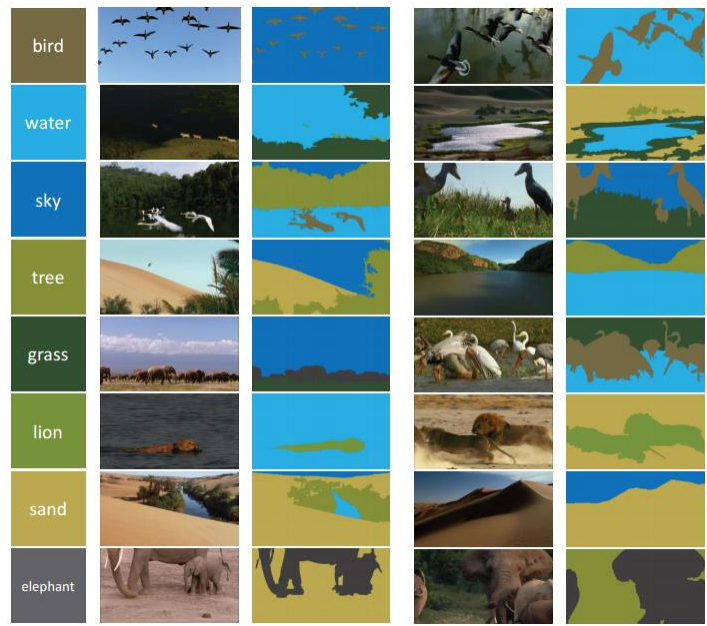

*Figure 3 Sample input frames of Wild8 Dataset and it's ground-truth*

### 2.1.4. YouTube-Objects Dataset

The YouTube-Objects (YTO) dataset consists of 126 videos collected from YouTube by querying for the names of 10 object classes [18]. It contains between 9 and 24 videos for each class and the duration of each video varies between 30 seconds to 3 minutes. YTO is a weakly annotated dataset with the name of the object in the scene and bounding box for 1 in every 10 frames. Each video in this dataset belongs to only one out of 10 classes and only relevant objects can be presented in that video. The collection contains 720000 frames that 6975 No. have been labeled and bounding box of them is also specified. Some samples of YTO dataset have been illustrated in Figure 4.

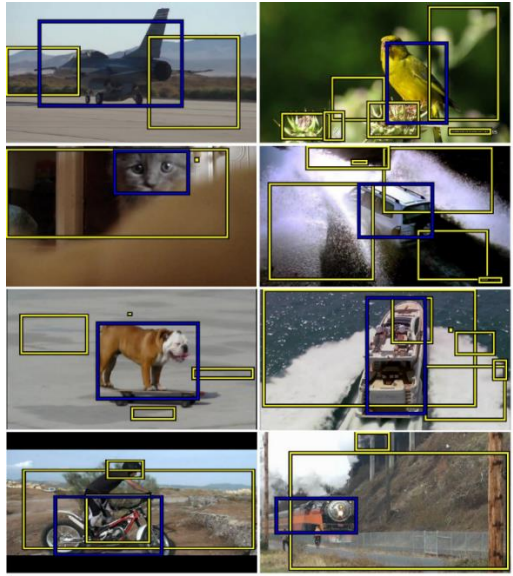

*Figure 4 Sample input frames of YTO Dataset*

### 2.1.5. SUNY Dataset

The SYNU dataset is a collection of Xigh.org videos [19] that consist of 8 videos with 24 classes. This dataset is challenging because there is a large number of classes in a small number of sequences. Also, some classes only exist in one sequence [17]. However, in spite of these challenges, the use of pixel-level labels in its frames causes to train as simpler than the weakly labeled datasets.

### 2.1.6. VSB100 Dataset

Galasso et al. provide ground-truth annotations for the Berkeley Video Dataset, which consists of 100 HD quality videos divided into train and test folders containing 40 and 60 videos, respectively [20, 21]. Each video was annotated by four different persons to decrease inaccurate labeling probability. Some samples of VSB100 dataset have been illustrated in Figure 5.

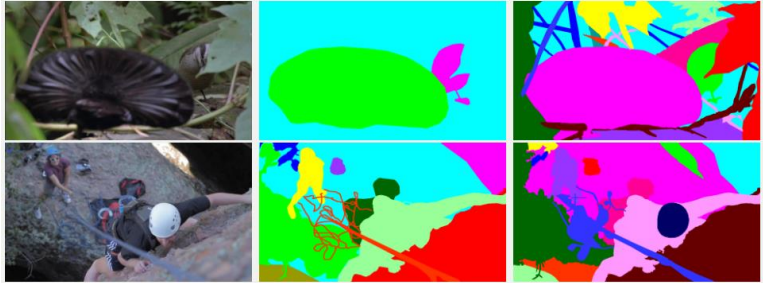

*Figure 5 Sample input frames of VSB100 Dataset and it's ground-truth that annotated by different persons*

### 2.1.7. SegTrack V2 Dataset

SegTrack v2 is a video segmentation dataset with full pixel-level annotations on multiple objects at each frame within each video [22]. This dataset consists of 14 videos with single object or interacting objects presented in each video. Unlike the most of the datasets in this area that a subset of dataset is selected and labeled as pixel-level, or all frames are weakly labeled, in this collection, all frames of all videos have labeled as pixel-level. SegTrack V2 dataset contains challenging cases including foreground/background occlusion, large shape deformation and camera motion. An example of inputs with different ground-truths are shown in Figure 6.

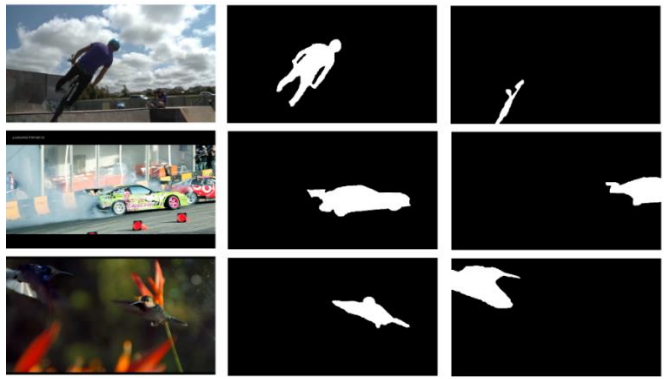

*Figure 6 Sample input frames of SegTrack v2 Dataset and it's different ground-truth*

### 2.1.8. Cityscapes Dataset

This dataset is a collection of recorded videos from 50 different cities in different seasons and different times of the day. The collection includes 5,000 frames with detailed labels (Figure 7) and 20,000 frame contains approximately labels (Figure 8) that consist of 30 different classes [23]. Depth of entities in the scene are also included in this dataset.

This dataset has been introduced recently but based on the high accuracy of its labeling, it will become one of the main references for performance evaluation in image and video processing area.

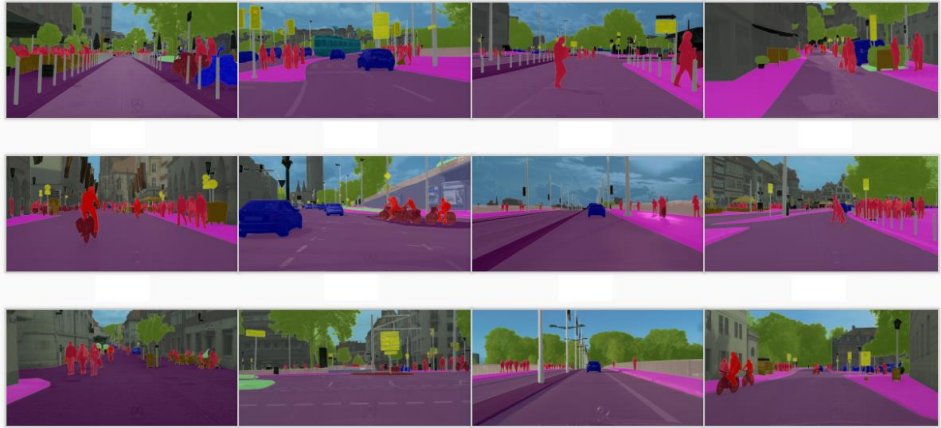

*Figure 7 Sample input frames of Cityscapes Dataset and it's detailed ground-truth*

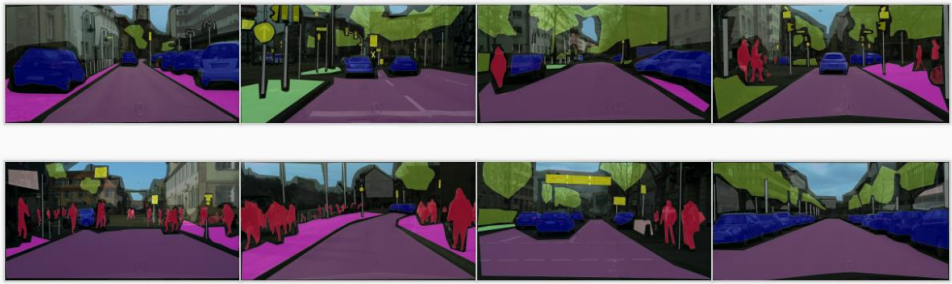

*Figure 8 Sample input frames of Cityscapes Dataset and it's approximately ground-truth*

There are some other datasets such as Daimler Urban Segmentation Dataset, Video Geometric Context, KITTI, LEUVEN, SIFT Flow and Stanford Background but a more detailed explanation of these items will be refused due to the lack of interest in recent years or based on their novelty and lack of similar comparable work.

## 2.2. Evaluation parameters

An important part in introducing a new system is to provide a full report on system performance, accuracy and precision in different conditions. Therefore, researchers attempt to compare their system results against others in order to prove that their system has a better performance than similar systems introduced by other researchers. The similar test conditions means being the same in testing datasets and evaluation parameters of quality. Consequently, over time, a number of test methods and parameters are being considered as a standard benchmark for comparing the methods with each other. Generally, evaluation parameters can be divided into two categories:

- ❖ The accuracy of labeling parameters
- ❖ The accuracy of segmentation parameters

Here, the most commonly measures that have been used for accuracy evaluation have been described including labeling performance and segmentation performance parameters.

### 2.2.1. Mean Accuracy per-pixel

PPA is the ratio of the correct pixels to the total pixels while ignoring the background [24]. It is a labeling performance parameter that is defined in Eq. 1. This measure sometimes has called as global accuracy.

$$1) \quad \sum_i n_{ii} \Big/ \sum_i t_i$$

Where $n_{ii}$ is the number of pixels of class i that is predicted correctly to belong to class i and $t_i$ is the total number of class i pixels.

### 2.2.2. Mean Accuracy per-class

MCA is the mean of class wise pixel accuracy [24]. It is a labeling performance parameter that is defined in Eq. 2.

$$2) \quad \frac{1}{n_{cl}} * \sum_i n_{ii}/t_i$$

Where $n_{cl}$ is the number of classes.

### 2.2.3. Mean Intersection over Union

Mean IU is a segmentation performance parameter that measure the overlap of two objects by calculating the ratio of intersection/union [25]. This is popular since it penalizes both over-segmentation and under-segmentation separately [24]. It is defined in Eq. 3.

$$3) \quad \frac{1}{n_{cl}} * \sum_i n_{ii} \Big/ t_i + \sum_j n_{ji} - n_{ii}$$

This parameter is defined as TP/(TP+FP+FN) where TP stands for true positive, FP means false positive and FN refers to false negative. However, for calculating the average of intersection/union score, it should calculate this score for the binary attribute and then divide it by the number of attribute (i.e. classes) [26].

### 2.2.4. Frequency Weighted IU

This Parameter is defined in Eq. 4.

$$4) \quad (\sum_k t_k)^{-1} * \sum_i t_i n_{ii} \Big/ t_i + \sum_j n_{ji} - n_{ii}$$

Since PPA parameter is biased in favor of large classes and discount small classes in datasets, researchers have interested on new consistency parameter that has been explained in next subsection.

### 2.2.5. Temporal Consistency

This parameter is based on consistency of tracks. A track is labeled consistent if all pixels along the track have the same label [27]. Therefore, the consistency of labeling is the fraction of tracks that are labeled consistent.

## 3. Recent Approaches

A wide range of researches have been done in the field of semantic segmentation. Besides the advantages of some methods over the other ones, they can be studied based on the type of data they accept as system input, the methods used to extract features from input data and the method of modeling and classification. In the following, we will describe these items.

### 3.1. Input of Semantic Segmentation Systems

A variety of inputs can be defined for segmentation systems. Binary inputs such as foreground and background, which are used to segment an object as a foreground and the rest of scene as a background, are among the simplest type of inputs [28, 29]. One of the most prominent definable applications for these systems is an anomaly detection system in a scene which models an anomaly as a category and defines the rest of the scene as an outlier category [30]. On the contrary, one can study several methods which are based on multi-class inputs [17, 19, 31-34]. In these systems, according to the categories existed in training dataset, multi-class outputs can be received from experimental scenes. The number of trained categories and the quality and accuracy of them depend on the number and quality of available categories in the training datasets of system.

Given the quality of available datasets, almost all the studies in this field in recent decade, have used color videos (RGB) [4, 5, 7, 17, 35, 36]. But in the meantime, a number of approaches have focused on 3D inputs [37-39]. Due to the recent features provided in some training datasets, some researchers have been attracted to the use of geographical coordinates on the scenes. In such systems, along with the semantic segmentation of the scene, the accuracy of segmentation can be increased with the help of geographic coordinates of different parts, and also appropriate geographic and place names can be used on decomposed parts of the scene [40].

### 3.2. Feature Extraction

Feature selection and extraction is one of the most influential steps in the design and implementation of machine learning systems. Video semantic segmentation is not separate from this principle. Selecting the best set of extractable features from training videos to create the best model is a major step in increasing the system efficiency and segmentation quality. The features used in different systems can be divided into two general categories. The first category includes the features extracted from the whole images, such as color and histogram features that will be discussed later. In contrast, there are region-based feature extraction techniques in which some regions of image are selected for feature extraction, then the features of each region are extracted and placed in feature space as a vector, separately. That's why the method of selection and extraction of image regions is important. This will be discussed in the following.

### 3.2.1. Super-voxel

As the term pixel that is used to refer to the smallest component of an image, the term voxel is used to refer to the smallest component of a video which can be considered as a 3D structure of images. This concept is remarkable and important, because, in many works introduced in the field of video processing, the researchers have initially done preprocessing on the input video in order to detect voxels and super-voxels, and then extracted the defined features from the super-voxels [17]. Various methods can be implemented to extract super-voxels, including mean shift, graph-based, hierarchical graph-based and SWA. According to the studies carried out, different extraction methods have shown different performance based on the type of dataset used and the desired application of video system designers [41]. This has caused the method of super-voxels extraction to be one of the most influential steps in the super-voxels based approaches.

### 3.2.2. Hand-craft Features

Defining new hand-craft features is one of the dynamic field after the emergence of a new research field of machine learning. After the creation of a new research field and the definition of its basic infrastructures, basic definable features in that field are provided by the founders, and over time, due to the increasing attention of researchers and the extension of the field among the academic community, the defined features are also further evolved and the quality parameters of the features such as comprehensiveness and separability are improved.

One of the fundamental features, which have been used in the majority of video works and semantic segmentation, are pixel color features of video frames [5, 7, 17, 32, 38, 42-44] which includes three features for RGB, three features for HSV, the statistical features obtained from the histogram and a number of histogram equalization methods.

Among the other set of features defined on each image pixel, we can mention to histogram of oriented gradient (HOG) which can be defined and implemented on both x and y axes [7, 32, 35, 45]. Of course, other types of histogram such as hue Color histogram and texton histogram are definable and extractable in video works [4, 33, 46].

Another set of hand-craft features which can defined on videos are appearance based features, such as across boundary appearance features, texture features and spatio-temporal appearance features, that some of them have been used in recent works [5, 7, 17, 32, 34-36, 44, 47].

Also, three-dimensional space features and 3D optical flow features have been used in the works done on RGBD datasets [37, 38].

### 3.2.3. Autonomic Features

In contrast to traditional methods using hand-craft features in designing of their systems, deep learning approaches have been proposed recently so that one of their features is to define and extract the best set of separable features from the datasets.

Because the convolutional deep networks have shown a very good performance in image and video processing tasks, a great interest have been observed in the use of these networks to define and extract features. Using of pre-trained Models on convolutional networks is one of the common methods for extracting automatic features from input data [8, 39, 48-55].

## 3.3. Modeling and Classification

The method used to model and learn different categories in a training dataset have a direct effect on classification accuracy and performance of a video semantic segmentation system. Therefore, various methods proposed in the field of machine learning have been used by many researchers, and each team have tried to improve their method and express its advantages. In the following, we will review the most common methods used for modeling and training. Given that these methods are shared between image and video semantic segmentation systems and can be used for both of them, we will also refer to some new works proposed in image semantic segmentation field.

### 3.3.1. Unsupervised Methods

Unsupervised methods have their own advantages and disadvantages. One of the advantages is that these methods doesn't need to dataset labeling, resulting in decreased costs. Also due to not being dependent on a particular dataset, unsupervised based systems can be considered more versatile than other existing systems. In contrast, the computational cost of these methods, particularly in the field of video, is very large, and some ways should be taken to control and reduce the computational cost and hardware requirements. Another disadvantage of these methods, which is within the scope of semantic segmentation concept, is that they cannot implement semantic segmentation without the use of labels. In other words, the semantic methods look for learning information from the classes they must segment, but unsupervised methods look for finding compatible regions in the scene. That is why these methods can be used as auxiliary methods along with other learning and modeling methods, so that the system segmentation can be provided with the help of labels and available semantic information in the scene. In the following, some of the most widely used algorithms in unsupervised methods will be mentioned.

A. Clustering Algorithms

K-means and mean-shift algorithms can be considered as the most popular clustering algorithms [3]. K-means algorithm looks for the best location for the centroid of k categories considered by user, and therefore it first randomly puts the points in features space and then, gradually select a better location for them to finish the algorithm. In contrast, mean-shift algorithm finds the cluster centers based on their maximum density-weighted in the feature space and in a similar way to hill-climbing algorithm with steps at determined interval, without need to define the number of clusters.

B. Graph-Based Algorithms

In image processing, these algorithms are used to present tree structure and the communications between image pixels, where the pixels are presented as nodes, and similarities and compatibility between them are displayed as connection weight between the nodes. In the video processing tasks, voxel is displayed as node. Performance of graph-based algorithms is based on obtaining the minimum possible spanning tree for initial structure drawn in the scene. This task is done according to the defined weight of edges or threshold in the system and the algorithms considered for scrolling minimum spanning tree. Also, graph-based algorithms are often used to find super-voxels in order to provide a precise definition of parts in scene and the relations between them [5-7, 37, 42].

C. Random Walk

Random walk algorithms are used in graph-based systems. The structure of these algorithms is such that different points of objects in the scene are selected, and then the probability of each point to reach to other points is calculated according to the gradient at that point. During classification, each pixel is assigned to the class with maximum probability of reaching. These algorithms can be implemented as iterative and non-iterative [40, 55, 56].

### 3.3.2. Support Vector Machine

Support vector machine, known briefly as SVM, is one of the most optimal classifiers for various problems with limited number of categories. These classifiers seeks to provide an optimal separator which has the longest average distance from all categories of feature space. Of course, when full separation of categories is not possible due to data disturbance or interactions between the features of different categories, the newer type of SVM with soft/hard margin can help the system designers. In such cases, kernel SVMs can also be used, which can provide higher separability for the set of available features through converting the feature space to a new space according to their kernel (Gaussian kernel, Sigmoid Kernel etc.) [57].

The initial version of support vector machine was based on binary classification and then, the "one vs one" and "one vs all" methods were introduced to categorize the problems including multiple categories. According to different types of inputs such as binary and multiple inputs as mentioned in Section 3.1, these methods can be used for modeling and classification [46, 47, 58].

### 3.3.3. Random Decision Forest

One of the proposed methods for learning models is ensemble learning in which a group of different classifiers are trained and the output of the trained model is obtained from composition of their group decision. Various methods have been introduced for ensemble learning. For example, all feature space may be trained for all models and consequently the output of all models can be used for the final decision of system, with the help of voting.

In a random decision forest (RDF) algorithm, usually random subspaces is used to divide the space needed for learning each of the classifiers and each of the models are trained on a subset of feature space. Because the trained models on these subspaces are decision trees, these methods are known as random decision forest [59].

If the trees are well trained on the dedicated subspaces of feature space, they will have a competitive power and accuracy of classification in comparison with other methods such as SVM and neural network. Due to their high speed in training and classification, RDFs are suitable options for video semantic segmentation [34, 44, 60-62].

### 3.3.4. Markov Random Field

Statistical models are classified as a general category within training models. Markov random field (MRF) is a subset of statistical models, which presents the relationship between random variables with an undirected graph. In these models, according to given possible values of all neighbors of a variable, the variable is independent from all other variables which are not on its neighborhood.

MRF model, sometimes known as undirected graphical model, has the same structure as Bayesian networks, and the difference is that the MRF network is undirected and supports circular connections. For this reason, the MRF has the ability to present and support dependencies in problems which cannot be modeled by Bayesian networks. Considering this point, the MRF structure have been used in a number of works which have been modeled with the help of statistical graphical models [24, 54, 62-64].

### 3.3.5. Conditional Random Field

Conditional random field (CRF) is another type of undirected statistical models, which is usually used in pattern recognition and the problems with structured prediction. In these problems, the classifier can apply its favorite label on input sample regardless of the label of neighboring samples. In machine vision tasks, CRFs are usually used to identify objects and segment the scene [1, 32, 35, 36, 51, 52, 65].

### 3.3.6. Neural Networks

It has passed over several decades since the construction of the first artificial neural network similar to the human nervous system, and over time, the structures introduced by the researchers have dramatically evolved. In the beginning, neural networks were composed of a single perceptron structure [66]. Over time, the perceptron's relations were defined, and given that computers' processing ability had also increased, the researches on various architectures of neural networks for various applications were highly expanded.

Today, different architectures of artificial neural networks have been provided, which can be distinguished based on connections between nodes of a layer, number of hidden layers, type of connections between layers, activation functions, etc. Considering the fact that deep neural networks have brought a new and revolutionary approach with themselves, neural networks are divided into two categories: traditional and deep, and the application of each category in video semantic segmentation is investigated.

#### A. Traditional Neural Networks

The use of traditional neural networks is one of the most common options for modeling in machine learning systems. Also, in video segmentation, after defining the desired features of system designers, with the help of neural networks and due to the fact that any complex feature space can be modeled with the increase of hidden layers, different classifiers have been introduced for these systems which, according to their layered architecture and activation functions, have some advantages over each other [43, 67-70].

B. Deep Neural Networks

Deep Neural Network (DNN) is a new field of neural networks, which recently provided a lot of progress on a variety of machine learning related topics. Several problems such as gradient descent in deep layers and back-propagation of neural network errors in different layers of traditional architecture, have existed in deepening the traditional neural networks. With the emergence of the new field of machine learning called deep learning which have proposed a new definition for deep neural networks, these problems have been solved and the ability to use deep networks with different capabilities have been provided for researchers.

Various structures have been proposed for DNNs, among them convolutional networks, recurrent networks and deep belief networks are the most widely used structures. Each of the mentioned structures have more progress than others in some topics of machine learning. According to the capabilities that are available in convolutional networks and the results provided by convolutional network based systems in the field of image and video processing, this type of network have become one of the most ideal and known options available for modeling [2, 8, 10, 39, 48, 49, 71].

Recently, fully convolutional networks (FCNs) have become one of the topics that has attracted much attention. FCN idea is to develop structure of convolutional networks (ConvNets) in order to support arbitrary input size [71]. At the beginning, this structure was used for one-dimensional and two-dimensional inputs [72, 73]. But over time, it was used for other applications in machine vision field such as image restoration, sliding window detection, depth estimation, boundary prediction and semantic segmentation. The researchers convert ConvNet to FCN by removing the last classifier layer of ConvNet network and converting all fully connected layers to convolutional layers, and thus, an end-to-end module is obtained for their defined machine vision problem [2, 8, 10, 39, 48, 71]. As mentioned in section 3.2.3, recently many researchers have used the convolutional networks as feature extractor [8, 39, 48, 49]. In the newest studies, a new convolutional module have been introduced by [74], which is specifically designed and optimized to predict the depth in the scene. In this module, a dilated convolution layer with the ability to integrate multi-scale conceptual information is implemented, which has achieved good results in video semantic segmentation. However, [27] was able to achieve better results with the help of CRF graphical model through coinciding pixels with a Euclidean distance space and using dilated convolution model trained in [74]. Furthermore, a new spatio-temporal module has introduced recently that consist of a FCN network and LSTM module as an end-to-end architecture [75] that achieved better results than [74] without the computational overhead of optical flow that have used in [27].

### 3.4. Limitation and Disadvantages

Most systems introduced in video semantic segmentation field have rarely used temporal features and temporal correlations between video frames, and in some cases, these features have been used but with some constraints and limitations included in definition of time window and temporal features [35, 36, 42, 45, 46, 62, 64]. For example, in [35], researchers have tried to establish temporal consistency relationships between frames with the help of objects' motion in them, and, through segmentation of objects and motion relations in a sequence, have proposed a method for semantic segmentation of video, which uses a spatio-temporal graphical model CRF and a labeling based on selecting the main frame of input and extracting its super-voxels. Also according to the graph-based method used in [42] for segmentation, the random selection method have been used to sparse the extracted graph so that high computational power will not be needed for video segmentation, and also, the frames' optical flow have been used to add motion- temporal information to the graph.

Some Limitations for building relationships between frames and using them can also be observed in recent works done by using deep convolutional networks [4, 5, 7, 17, 32, 37, 49, 53]. For example, in [4], the shape compatibility of the object traced in time intervals have been used for temporal compatibility between frames. In the methods that have used graph structures for segmentation, temporal neighbors have been also considered in addition to spatial neighbors relations in the graph [5, 7, 17]. Of course, along with the above-motioned approach, [7] have used spatio-temporal appearance features, spatio-temporal motion features and the similarity of the shapes of objects at different times as system features. In order to implement super-voxels in graph-based works, spatio-temporal super-voxels including desired region in deferent frames can be defined, and the relationships between regions in graph may be defined by using average relationships in different frames [17]. In addition, the optical flow have been implemented in recent approaches which are based on convolutional networks as a temporal feature [27, 37].

Another disadvantage of motioned systems is not-being end-to-end, which is a major weakness in design of such systems. Of course, some recently introduced systems with deep networks use both spatial and temporal features as an integrated system, but due to fixed temporal window, they have shown a poor adaptability to different inputs [27, 39].

### 3.5. Recent Comparison

Given that CamVid and NYUDv2 are among the most common datasets used to evaluate the efficiency of video semantic segmentation systems, the latest results reported on mentioned datasets are provided in Table 1 and Table 2, respectively.

As shown in Table 1, the high growth observed in system [76] is due to using of FCN network in system architecture, training the network by using different input sizes and then merging the features extracted from them, so that a multi-scale model is obtained by using this method. But, the dilated Convolutional network introduced in [74] have been able to achieve higher performance in semantic segmentation of CamVid dataset based on its ability to integrate multi-scale information for larger size inputs and providing a better segmentation by better depth prediction. Also, the spatio-temporal architecture introduced in [75] have increased the dilated8 model performance due to the use of temporal features with LSTM embedded modules. The method presented in [27] have improved the accuracy of segmentation compared with other methods, considering the improvements which were applied within the features space and the Euclidean distances between them.

Table 1 A comparison of recent approaches proposed for semantic segmentation on CamVid dataset

| Mean IU | Class Acc | Pixel Acc | |
|---|---|---|---|
| - | 62.5 | 83.9 | Tighe and Lazwbnik [77] |
| - | 72.47 | 76.35 | Ravi, et al. [44] |
| 53.6 | - | - | ALE [78] |
| 42.0 | 51.2 | 83.3 | SuperParsing [79] |
| 47.2 | 62.4 | 82.8 | Liu and He [32] |
| 46.4 | 62.9 | 84.3 | SegNet [48] |
| 45.8 | 62.5 | 82.5 | Liu, et al. [35] |
| 61.6 | - | - | DeepLab-LFOV [76] |
| 65.3 | - | - | Dilation8 [74] |
| 66.12 | - | - | Kunda, et al. [27] |
| 65.9 | | | Fayyaz, et al. [75] |

As shown in Table 2, the use of deep convolutional network with pre-trained models for feature extraction and adding the temporal relationships between frames with the help of optical flows and relationships of different regions in the scene, provide a high performance for semantic segmentation of NYUD dataset [39]. In order to use semantic data contained in patches and backgrounds optimally, designers of the system presented in [52] have achieved a better performance in segmenting NYUD dataset by using a combination of CNN and CRF and multi-scale networks to obtain semantic relationships in patches and extract semantic relationships in backgrounds.

Table 2 A comparison of recent approaches proposed for semantic segmentation on NYU-V2 dataset

| Mean IU | Class Acc | Pixel Acc | |
|---|---|---|---|
| 34.1 | 45.1 | 65.6 | Eigen and Fergus [53] |
| 40.6 | 53.6 | 70.0 | Lin, et al. [52] |
| 28.6 | - | 60.3 | Gupta, et al. [60] |
| 34.0 | 46.1 | 65.4 | FCN [71] |
| 40.1 | 53.8 | 70.1 | He, et al. [39] |
| - | 41.0 | 50.5 | SegNet [48] |
| 32.8 | 44.9 | 64.3 | Fayyaz, et al. [75] |

## 4. Convolutional Neural Networks Methodology

Convolutional neural networks (CNN) have been used in a wide range of problems in the field of machine learning. These networks have shown a very high capability in machine vision problems. Through changing the structure of the CNN networks, some researchers were able to introduce new architectures which have better performance in different problems. Recently, a new architecture of CNN networks, known as fully convolutional network (FCN), was introduced. In this network, the classification layer at the end of the CNN structure has been removed and fully connected layers have been converted to convolutional layers.

Input structure of each layer in a convolutional network is h*w*d, where h, w and d are width, length and depth of input, respectively. According to the connectivity structure of network, location in the upper layers of network is linked to the first location in input image, where the motioned region is called receptive filed.

The primary components of a convolutional network are convolution operations, pooling operations and different activation functions, which are applied on input local regions and are dependent on the input relative coordinates. If $x_{ij}$ is input array of a layer for coordinates (i.j), and $y_{ij}$ is defined as output, then the output can be calculated by eq. 5 [71]:

5) $$y_{ij} = f_{ks}(\{X_{si+\delta i, sj+\delta j}\}\ 0 \leq \delta i, \delta j \leq k)$$

where, $k$ is defined as the size of the kernel used, $s$ is stride and $f_{ks}$ is type of the layer which can be of various types such as convolution coefficient matrix, mean or min or max pooling layer, non-linear activation function and so on.

The structure of equation 5 obeys the transformation law for kernel size and stride, defined by Eq. 6:

6) $$f_{ks} \circ g_{k's'} = (f \circ g)_{k'+(k-1)s', ss'}$$

Deep networks calculate and generate a general nonlinear function, but in networks with such defined layer, a non-linear filter is calculated and generated which can be called FCN. A FCN network can be used on input with any size and generate output with desired size.

## 5. Challenges

Recently, semantic segmentation systems have achieved a maturity in overall. These approaches have shown a great performance in evaluation exercises. Especially with recent progress in deep CNN approaches, almost all the reports on different datasets have passed by these methods. Although these records have made by recent approaches, still some problems have remained in this field.

Recent approaches have focused on segmenting small as well as detailed objects. Each of them have reported several achievements on this objects but still there is a lot of space for progress. Typically, a large percentage of errors are related to these objects. Therefore, this challenge can be one of the topics that researchers can choose as future work.

Some problems in video datasets have become to permanent challenges for researchers. These problems can be explained as follows:

- ❖ Lens flare or vignetting: This is the effect of light that is scattered in lens or the frame image that is darker in some corners.
- ❖ Blurriness: This problem can occurs in some cases such as the camera has a wrong focus, object movements, smoke, etc. The blurred object has become hard to detect with details and segment.
- ❖ Partial occlusions: This problem occurs when some part of object have blocked by other objects. In this situation, segmenting a blocked object has become a challenge for segmentation system.
- ❖ Coverage: This is the situation that commonly occurs in wildlife documentaries and turn the object detection and segmentation into a challenging task.
- ❖ Point of view: Each approach has a trained model that use for classification in test phase. These models must have a huge set of training data to achieve a full description of objects. In some cases, if objects from a special view point has not delivered for training, the system has become vulnerable in test phase.

## 6. Future Work

Semantic segmentation is a growing research field. We surveyed this scope and studied state of the art methods in this domain. Although, researchers have proposed good appropriate methods, they are not adequate. However, there are case studies that need more attention. In the following, we have mentioned some of these cases:

- ❖ Temporal Features: Recent approaches are mainly designed for image semantic segmentation, while there is a growing thirst for video semantic segmentation. In video, frames are correlated to each other and have temporal data. There is not enough focus on this problem in recent works. Proposing more spatio-temporal semantic segmentation systems is needed in this scope.
- ❖ Objects semantic relations: In semantic segmentation problems, there are several semantic relations inside a scene. Not only each object alone has a semantic identity, but also has a subset of objects. In other words, an object has a solo semantic identity, and a group semantic identity. The group semantic identity comes from the neighborhood objects. The group identity affects the solo semantic identity and vice-versa. Studying these relations and their effects is a good case study for future researches.

## 7. Conclusion

This paper has studied recent approaches that have been proposed in semantic segmentation field that is one of the main tasks in many computer vision problems. In the first section, some introduction has explained to introduce this research area. In the next section, several common datasets have explained that have been used for video segmentation. In addition, the most commonly used parameters in semantic segmentation evaluation have expressed. In section 3, a complete brief review has performed on a variety of recent researches that have done in this field. This section consists of two parts that have focused on features selection and segmentation methodology in recent approaches. Section 4 explained the fundamental methodology have used in CNN and FCN based approaches. At the end, a number of important challenges and future works have explained to clear the future research path for researchers.